\documentclass{article}

\usepackage[final, nonatbib]{neurips_2020}

\usepackage[utf8]{inputenc} 
\usepackage[T1]{fontenc}    
\usepackage{hyperref}       
\usepackage{url}            
\usepackage{booktabs}       
\usepackage{amsfonts}       
\usepackage{nicefrac}       
\usepackage{microtype}      

\usepackage{xcolor}
\usepackage{caption}
\usepackage{subcaption}
\usepackage{graphicx}

\usepackage{amssymb}
\usepackage{amsmath}
\usepackage{amsthm}
\theoremstyle{definition}
\newtheorem{definition}{Definition}[section]

\usepackage{mathrsfs}
\usepackage{xcolor}

\title{Inductive Bias and Language Expressivity \\ in Emergent Communication}

\author{%
  Shangmin Guo\\
  School of Informatics\\
  University of Edinburgh\\
  \texttt{s.guo@ed.ac.uk} \\
  \And
  Yi Ren\\
  Department of Computer Science\\
  University of British Columbia\\
  \texttt{renyi.joshua@gmail.com}
  \And
  Agnieszka Słowik\\
  Department of Computer Science and Technology\\
  University of Cambridge\\
  \texttt{agnieszka.slowik@cl.cam.ac.uk}
  \And
  Kory Mathewson\\
  DeepMind\\
  \texttt{korymath@google.com}\\
  \And

}

\begin{document}

\maketitle

\begin{abstract}
Referential games and reconstruction games are the most common game types for studying emergent languages. We investigate how the type of the language game affects the emergent language in terms of: i) language compositionality and ii) transfer of an emergent language to a task different from its origin, which we refer to as language expressivity. With empirical experiments on a handcrafted symbolic dataset, we show that languages emerged from different games have different compositionality and further different expressivity.
\end{abstract}

\section{Introduction}

With the development of deep reinforcement learning (DRL) techniques \cite{silver2016mastering} and its application in natural language understanding (NLU) \cite{narasimhan2015language}, recent works \cite{lazaridou2017multi, lazaridou2018iclr, ivan2017nips, lee2017emergent, li2019ease} pursue another feasible method for NLU, i.e. to facilitate human-like communication based on interactions between agents in either physical or virtual worlds.
To be more specific, the virtual worlds are implemented by games which are usually one of the following two types: i) referential games (which is a variant of the signalling game proposed by \cite{Lewis_refgame}) \cite{lazaridou2017multi, ivan2017nips, li2019ease, guo2019emergence, ren2020compositional, chaabouni2020compositionality}; and, ii) reconstruction games (in which the task for agents is to reconstruct the original input by communication) \cite{guo2019dissertation, chaabouni2020compositionality}.

Based on both types of games, lots of works focus on introducing inductive bias into the systems to make the emergent communication protocols between agents to have more natural-language-like properties \cite{gupta2020analyzing, kottur2017natural}.
Among all properties of natural languages, an important one is compositionality that helps us humans express novel concepts via structured combinations of simpler words and phrases \cite{ren2020compositional}. Thus lots of existing works focus on facilitating compositional languages in multi-agent communication \cite{ren2020compositional, kottur2017natural, cogswell2019emergence, mordatch2018emergence}.
None of the existing works, however, has compared the inductive bias inherently from the games.
Existing works \cite{hermann2017grounded, ren2020compositional} focus on a single type of language games: either referential \cite{hermann2017grounded} or reconstruction (e.g. in \cite{chaabouni2020compositionality}). 

In this work, on the other hand, we study the inductive bias provided by a particular task and its effect on language structure in the form of:

\begin{enumerate}
    \item \textbf{compositionality}: an property of languages that allow expressing novel concepts by compositing existing semantic units, and could be measured by certain metrics, e.g. topological similarity proposed by \cite{comp_measure01}; 
    \item \textbf{expressivity}: an indirect measurement of the information (about the inputs) contained in a language.  We experimentally measure to what extent an emergent language generalises to tasks that are different from the task where the language emerge, e.g. generalisation performance of emergent language from referential game on reconstruction game. 
\end{enumerate}

Intuitively, we are interested in reducing the cost and the noise of developing a new language for each task. Hence we usually expect an emergent language with the inductive bias we are interested in, and at the same time, with a high expressivity for different downstream tasks.

Through experiments on both referential and reconstruction games based on EGG, a Python package implemented by \cite{Kharitonov2019Egg}, we find that:

\begin{enumerate}
    \item game design has a significant influence on compositionality of emerged languages;
    \item emergent languages from different games have different expressivity.
\end{enumerate}

Our codes are released at \hyperlink{https://github.com/Shawn-Guo-CN/GameBias-EmeCom2020}{https://github.com/Shawn-Guo-CN/GameBias-EmeCom2020}.

\section{Background}

\begin{figure}[h]
\centering
\begin{subfigure}{.5\textwidth}
  \centering
  \includegraphics[width=.9\textwidth]{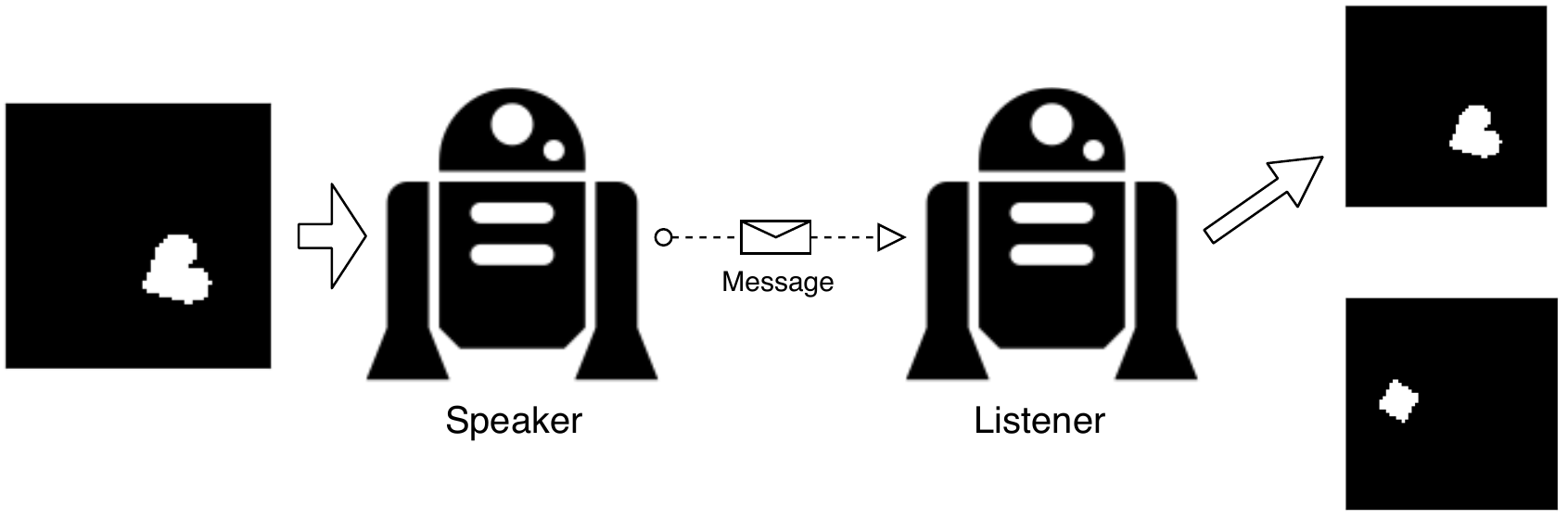}
  \caption{Referential}
  \label{fig:refer_game}
\end{subfigure}%
\begin{subfigure}{.5\textwidth}
  \centering
  \includegraphics[width=.9\textwidth]{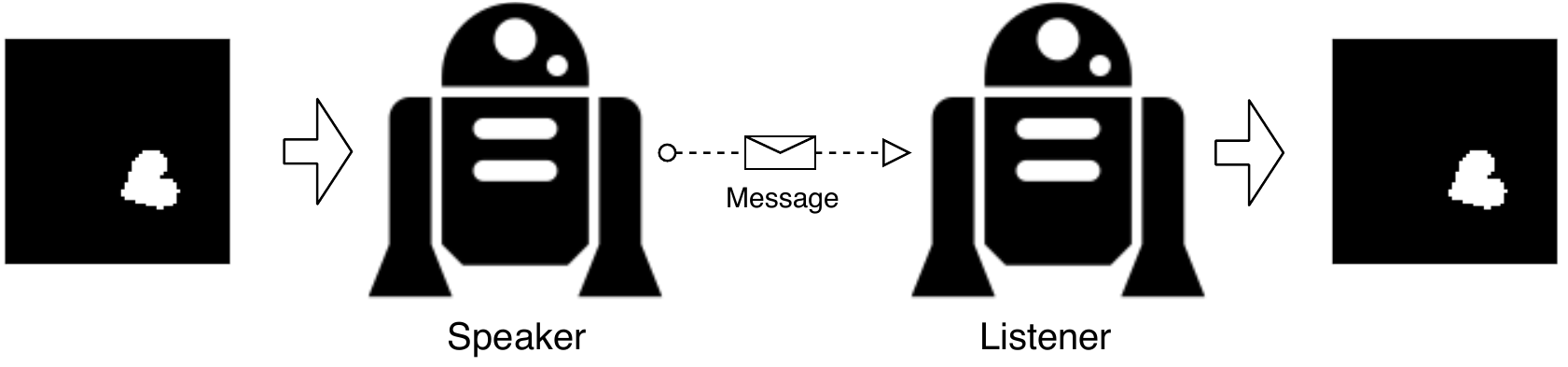}
  \caption{Reconstruction}
  \label{fig:recon_game}
\end{subfigure}
\caption{Two types of games used in our experiment. The sample input images are from \cite{dsprites17}.}
\label{fig:game_diagram}
\end{figure}

Following the assumption that ``language is to make things happen''\cite{wittgenstein2009philosophical}, more and more of natural language processing research turns to simulating emergent languages that share similar properties to our natural languages in virtual interactive environments \cite{lazaridou2017multi, lazaridou2018iclr, ivan2017nips, lee2017emergent}.
More specifically, to better align emergent languages with natural languages, many works focus on developing compositional languages in multi-agent games~\cite{kottur2017natural, mordatch2018emergence, guo2019emergence, ren2020compositional, slowik2020exploring, li2019ease}, based on a common assumption that compositional languages could be better generalised to the unseen samples.
Besides, some DRL models also take natural language instructions as auxiliary inputs to help improving the generalisability and zero-shot performance of DRL models \cite{luketina2019survey}.
However, \cite{guo2019dissertation} and \cite{chaabouni2020compositionality} both suggest that compositionality is not a necessary condition for generalisability.
Under some configurations of language games, \cite{guo2019emergence} even shows that compositional languages may have worse learnability than the ``naturally emergent language''.
Based on all the above results, we argue that the game settings actually could introduce their inherent inductive bias into the emergent languages and, further, they would affect the expressivity of the emergent languages\footnote{We will introduce more about the expressivity of emergent languages in Section~\ref{sec:expressivity}.}.
To measure it, we need to implement at least two types of game.
Thus, following the typical settings in existing works, we choose the following games in this work: i) referential game which is illustrated by Figure~\ref{fig:refer_game}; ii) reconstruction game which is illustrated by Figure~\ref{fig:recon_game}.

Besides, to avoid the complexity of image inputs, we hand crafted a dataset where each sample is a concatenation of one-hot vectors.
Suppose that we want to model $N_a$ attributes, and there are $N_v$ possible values for each attribute, we then could have $N_v^{N_a}$ different combinations, i.e. $N_v^{N_a}$ corresponding samples.
In our case, we set $N_a=3$ and $N_v=6$, thus $216$ samples in total, and they are randomly split into a training set (containing 173 samples) and a test set (containing 43 samples). Details about the full configuration of our games and models can be found in the Appendices.

\section{Compositionality}
\label{sec:compositionality}

With the help of compositionality, we could easily express novel or complex concepts with simpler semantics units we already acquired.
In other words, compositionality can help the agent generalise to some novel concepts (e.g., the non-seen concepts in test set) by combining the learned concepts from the training set.  \cite{smith2003complex}.
Further, we notice that the generalisation ability is also a concern of the researchers focusing on disentanglement representation learning, because a compositional representation would be helpful to the related but unseen tasks by disentangling each salient features to distinct dimensions \cite{chen2016infogan}.
Thus, we first investigate the influence of different language games on the compositionality of emergent languages, which we argue is an important property for emergent languages to pursue and could lead to an interesting interaction with disentangled representation learning \cite{chaabouni2020compositionality}. 

Although measuring compositionality is not a completely solved problem, there are some works proposing certain metrics for situations where true generative factors are accessible, e.g. \cite{comp_measure_tre} and \cite{comp_measure01}.
Considering the fact that there is no variance in emergent languages and the true generative factors of data samples are accessible, we use topological similarity proposed by \cite{comp_measure01} as the metric.
To be more specific, we take: i) edit distance for message space; ii) hamming distance for message space; iii) Pearson correlation coefficient for the correlation between distance in those two spaces.

\begin{figure}
    \centering
    \includegraphics[width=0.6\textwidth]{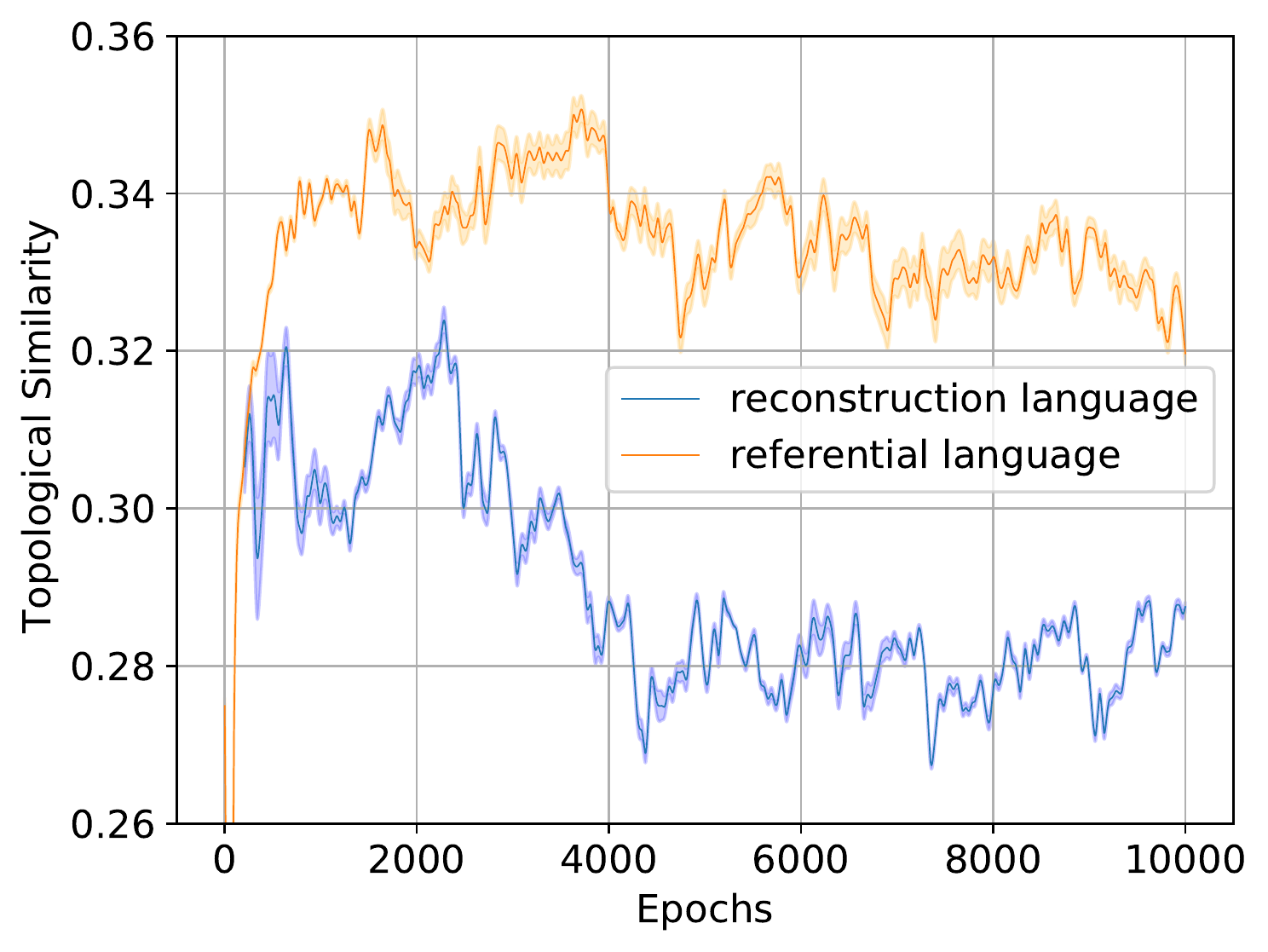}
    \caption{Topological similarity during training in reconstruction and referential games. The two lines are the average value from 10 runs, and the shadow area indicates the corresponding variance. Note that we run $10,000$ epochs on each game, as the Gumbel-softmax trick \cite{gumbel} suffers from a high variance problem with $\rho=1$ and we want to make the distribution of emitting symbols more close to one-hot categorical distribution. The diagram shows that the topological similarity of reconstruction language is higher than the referential language.}
    \label{fig:exp1}
\end{figure}

With ten different runs for each game type, the topological similarity curves in referential and reconstruction games are both drawn in Figure~\ref{fig:exp1}.
As we can see in the figure, the topological similarity of emergent languages from \textbf{referential game} is significantly higher than the one from \textbf{reconstruction game}, which could be further confirmed by a $t$-test between the two curves.
With $p = 1.55\times 10^{-29} \ll 0.01$, we conclude that the null hypothesis, the mean of converged topological similarity of emergent language from the two games is identical, can be rejected.

\section{Expressivity}
\label{sec:expressivity}






Inspired by \cite{smith2013linguistic} and the transfer learning from machine learning community, here we introduce an informal definition of the partial order of expressivity between different emergent languages. Suppose there are two emergent languages, $L_A$ and $L_B$, and the expressivity of them are $\mathcal{E}_{L_A}$ and $\mathcal{E}_{L_B}$ respectively. With a set of language games $\mathcal{G}$, we then informally define $\mathcal{E}_{L_A} \geq \mathcal{E}_{L_B}$ as follow:

\theoremstyle{definition}
\begin{definition}[$\mathcal{E}_{L_A} \geq \mathcal{E}_{L_B}$]
If the generalisation performance of $L_A$ is always as good as $L_B$ for all language games, i.e. $\mathcal{E}_{L_A}^{g} \geq \mathcal{E}_{L_B}^{g}  \forall g\in \mathcal{G}$, and the generalisation performance of $L_A$ is strictly better than $L_B$ on some language games, i.e.  $\exists g' \in \mathcal{G},\ s.t.\ \mathcal{E}_{L_A}^{g'} > \mathcal{E}_{L_B}^{g'}$, we then say the expressivity of $L_A$ is higher than $L_B$, i.e. $\mathcal{E}_{L_A} \geq \mathcal{E}_{L_B}$.
The metric of generalisation performance could be varied according to the types of games.
\end{definition}

\begin{figure}[h]
\centering
\begin{subfigure}{.5\textwidth}
  \centering
  \includegraphics[width=.93\textwidth]{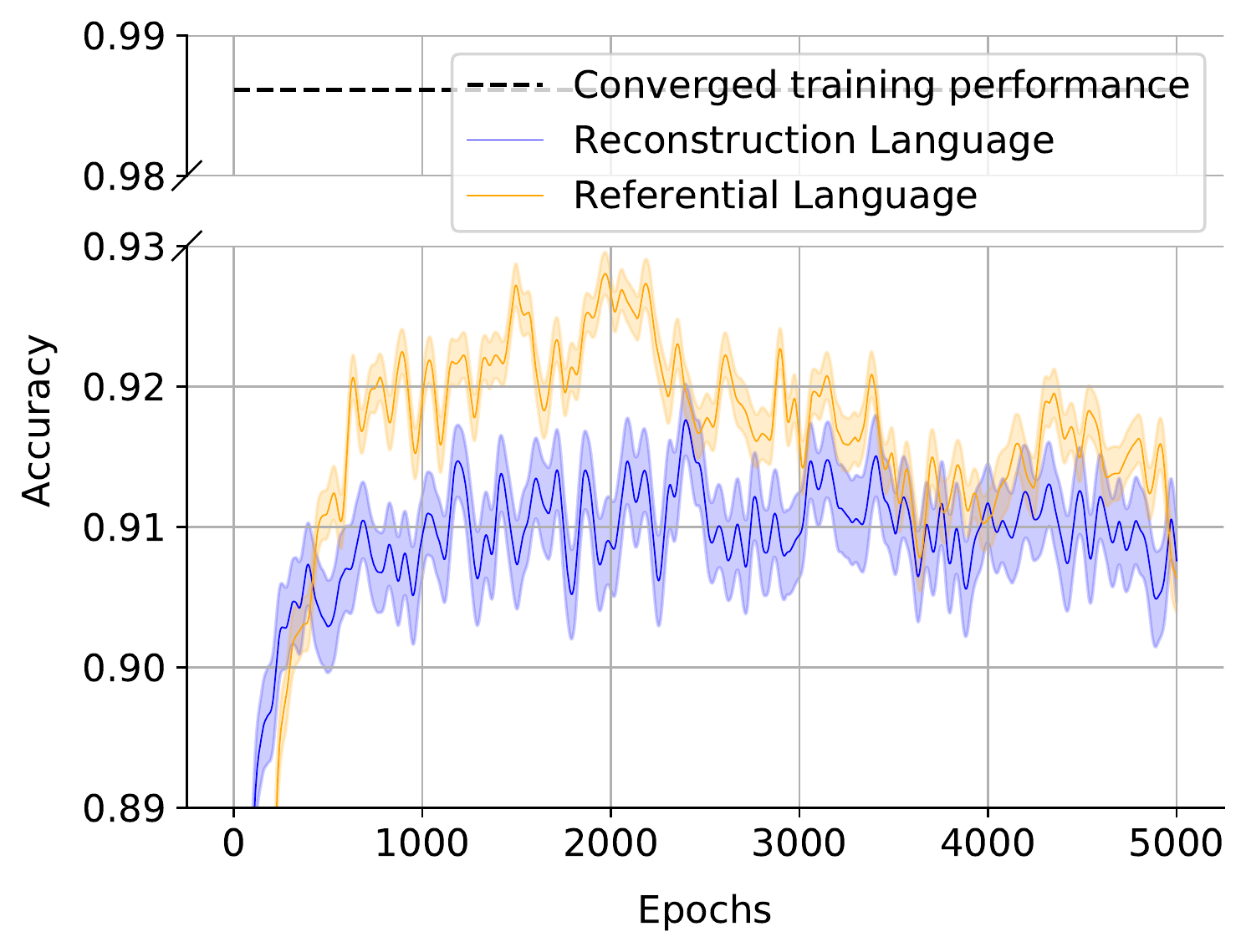}
  \caption{On referential game}
  \label{fig:exp2_refer_game}
\end{subfigure}%
\begin{subfigure}{.5\textwidth}
  \centering
  \includegraphics[width=.9\textwidth]{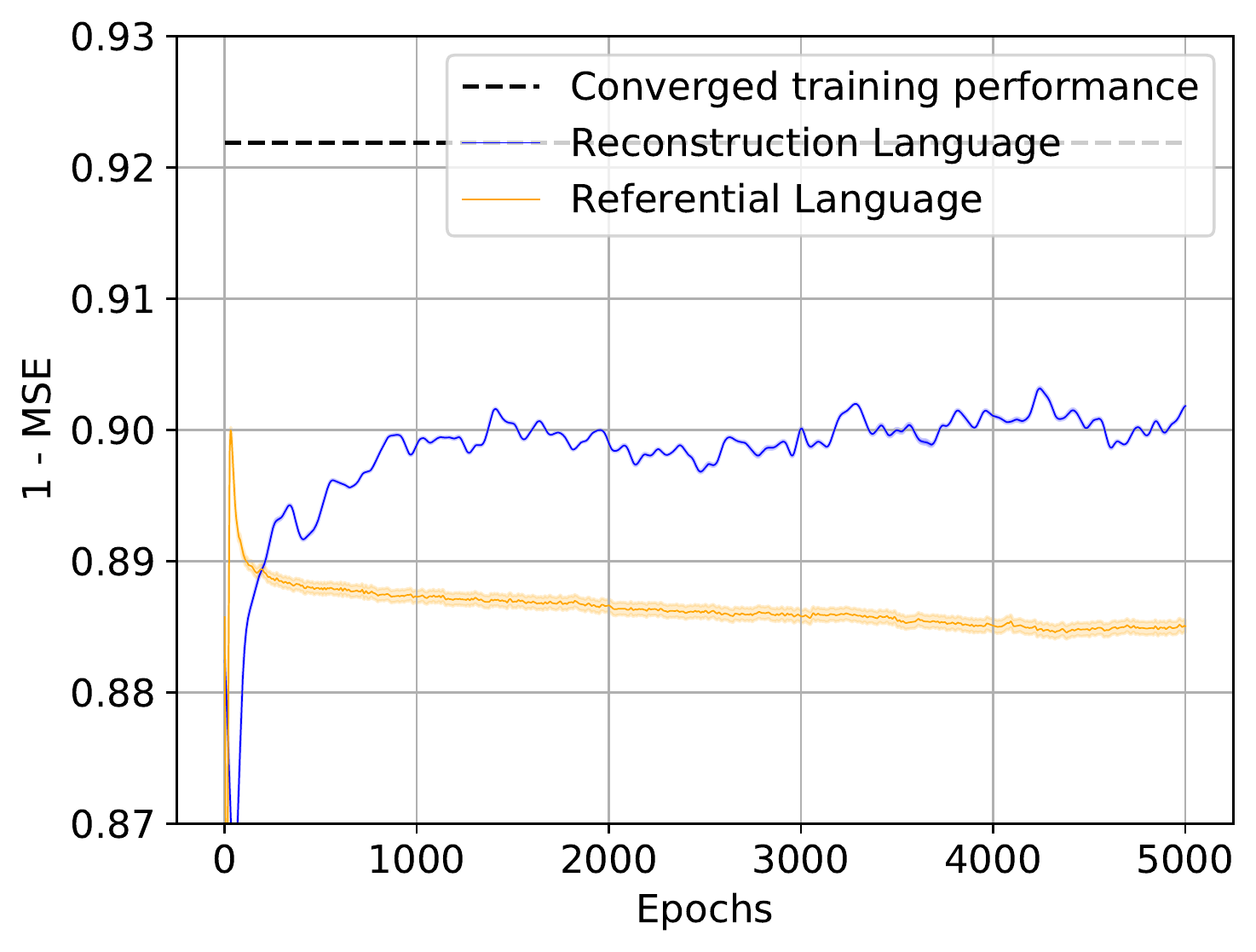}
  \caption{On reconstruction game}
  \label{fig:exp2_recon_game}
\end{subfigure}
\caption{Generalisation performance of different emergent languages on different types of games. Note that all the curves have been smoothed, and they are the averaged performance from 10 runs. The dotted lines are the converged training performance, and the shadow area are the corresponding variances. Besides, to make higher mean better in both figures, we converter the $MSE$ to $1 - MSE$.}
\label{fig:exp2}
\end{figure}

Take our settings in Section~\ref{sec:compositionality} for example, to compare the expressivity of emergent languages from referential ($L_{refer}$) and reconstruction ($L_{recon}$) game, we need to: 
i) split the whole dataset into training and test sets for both language games; 
ii) train agents on both of them to obtain $L_{refer}$ and $L_{recon}$; 
iii) train listening agents with only message-input pairs of $L_{refer}$ and $L_{recon}$ but without speaking agents\footnote{Thus the listening agents have to depend solely on the emergent language to complete the game.} on both games, and track the generalisation performance of $L_{refer}$ and $L_{recon}$; 
iv) on each game, do a $t$-test to see if there is a significant difference between the generalisation performance of the two languages.
Following the common settings in the community, we measure the generalisation performance on reconstruction and referential game by mean squared error (MSE) and accuracy on test dataset respectively.

Note that expressivity and mutual information are two different functions, as the inputs for expressivity are messages and tasks whereas the inputs for mutual information are input and messages.
Meanwhile, MI would monotonically increase and limits to infinity and thus become meaningless, as the emergent language would become more and more deterministic during the training\footnote{Emergent language is a mapping function from input space to message space, and it would become more and more deterministic such that listener could success in the game \cite{ren2020compositional}.}.

After running 10 times of the above procedure, we draw the generalisation performance of $L_{refer}$ (orange lines) and $L_{recon}$ (blue lines) on referential game and reconstruction game in Figure~\ref{fig:exp2_refer_game} and Figure~\ref{fig:exp2_recon_game} respectively.
As shown in Figure~\ref{fig:exp2_refer_game}, $L_{recon}$ obtains as good generalisation performance as $L_{recon}$ on referential game, which is indicated by that two curves overlap with each other, i.e $\mathcal{E}_{L_{recon}}^{refer} \approx \mathcal{E}_{L_{recon}}^{refer}$.
On the other hand, the generalisation performance of $L_{recon}$ is obviously better (higher) than the $L_{refer}$, i.e. $\mathcal{E}_{L_{recon}}^{recon} > \mathcal{E}_{L_{recon}}^{recon}$.

With a null hypothesis, ``the means of the generalisation performance are identical'', we obtain the following test results for experiments results demonstrated in Figure~\ref{fig:exp2}-(a) and (b) respectively: 
i) $p=0.741541>0.01$ on referential game, which means we cannot reject the null hypothesis; 
ii) $p=5.09859\times 10^{-64}\ll 0.01$ on reconstruction game, which means we can reject the null hypothesis.
The results indicate that both $\mathcal{E}_{L_{recon}}^{refer} \approx \mathcal{E}_{L_{recon}}^{refer}$ and $\mathcal{E}_{L_{recon}}^{recon} > \mathcal{E}_{L_{recon}}^{recon}$ hold.
Therefore, we could conclude that the expressivity of $L_{recon}$ is higher than that of $L_{refer}$, which also matches our intuition that the language used to reconstruct the original inputs should contain more information than the one used only to tell the difference between them.

\section{Future work and Conclusion}
We are presenting results of work in progress. We show how different types of games may introduce inductive bias to the emergent language.
For the referential game and the reconstruction game, which are two widely studied tasks in the emergent communication community, we compare their converged compositionality and expressivity.
The results show that the language emerged in a referential game is more compositional than that emerged in a reconstruction game, which we argue is caused by the target selection process inherently introducing a bias of distinguishing different attributes.
On the other hand, as the reconstruction game requires the agent to learn more details about the input signal, the expressivity of the emergent language from a reconstruction game is higher than the one emerged in referential game.
In summary, our work demonstrates that the game selection plays a crucial role in emergent communication study: a suitable game can let the emergent language contain sufficient useful information and have the desired properties. Future avenue of research is to repeat the experiments using complex inputs, such as images.

\clearpage

\bibliographystyle{plain}
\bibliography{neurips_2020}

\clearpage

\section*{Appendix A: Game Configurations}

The details about the configuration of our games are listed in Table~\ref{tab:game_config}.

\begin{table}[h!]
    \centering
    \begin{tabular}{l|c}
        \hline
        \multicolumn{1}{c|}{Description} & Value \\
        \hline
        Number of candidates in referential game & 2 (1 target and 1 distractor) \\
        Length of the communicating messages & 3 \\
        Number of symbols available for communication & 7 \\
        \hline
    \end{tabular}
    \caption{Configurations of our games.}
    \label{tab:game_config}
\end{table}

\section*{Appendix B: Model Configurations}

The details about the configuration of our models and the training procedure are listed in Table~\ref{tab:model_config}.

\begin{table}[h!]
    \centering
    \begin{tabular}{l|c}
        \hline
        \multicolumn{1}{c|}{Description} & Value \\
        \hline
        Batch size & 64\\
        Encoder for inputs & 3-layered multi-layer perceptron \\
        Size of hidden layer in encoder & 256\\
        Learning rate & $10^{-3}$ \\
        Temperature $\tau$ for Gumbel-softmax trick  & $1.0$ \\
        Optimisation algorithm & Adam\cite{kingma2014adam} \\
        \hline
    \end{tabular}
    \caption{Configurations of our models and the training procedure.}
    \label{tab:model_config}.
\end{table}

For all the other configurations not mentioned in Table~\ref{tab:game_config} and Table~\ref{tab:model_config}, we keep them as the default value in the EGG framework \cite{Kharitonov2019Egg}.

\end{document}